\def\year{2024}\relax
\documentclass[letterpaper]{article} %
\usepackage{aaai24}  %
\usepackage{times}  %
\usepackage{helvet}  %
\usepackage{courier}  %
\usepackage[hyphens]{url}  %
\usepackage{graphicx} %
\usepackage{natbib}  %
\usepackage{caption} %
\DeclareCaptionStyle{ruled}{labelfont=normalfont,labelsep=colon,strut=off} %
\usepackage{algorithm}
\usepackage{algorithmic}

\usepackage{soul}
\usepackage[dvipsnames]{xcolor}

\newcommand{\change}[2]{#2}

\usepackage{newfloat}
\usepackage{listings}
\floatstyle{ruled}
\newfloat{listing}{tb}{lst}{}
\floatname{listing}{Listing}
\title{Symbolic Numeric Planning with Patterns}

\author {
    Matteo Cardellini\equalcontrib\textsuperscript{\rm 1, 2},
    Enrico Giunchiglia\equalcontrib\textsuperscript{\rm 2},
    Marco Maratea\textsuperscript{\rm 3}
}
\affiliations {
    \textsuperscript{\rm 1} DAUIN, Politecnico di Torino, Italy\\
    \textsuperscript{\rm 2} DIBRIS, Università di Genova, Italy\\
    \textsuperscript{\rm 3} DeMaCS, Università della Calabria, Italy\\
    matteo.cardellini@polito.it, enrico.giunchiglia@unige.it, marco.maratea@unical.it
}

\usepackage{bibentry}

\usepackage{xspace}
\usepackage{amsmath}
\usepackage{amsfonts}
\usepackage{amssymb}
\usepackage{comment}
\usepackage{bm}
\usepackage{amsthm}
\usepackage{todonotes}

\newtheorem{theorem}{Theorem}

\newcommand{\set}[1]{\ensuremath{\{#1\}}\xspace}
\newcommand{\tuple}[1]{\ensuremath{\langle #1 \rangle}\xspace}

\newcommand{\op}[1]{\mathrm{#1}}

\newcommand{\ttt}{\texttt}

\newcommand{\pluseq}{\mathrel{+}=}
\newcommand{\minuseq}{\mathrel{-}=}

\newcommand{\mA}{\mathcal{A}\xspace}

\newcommand{\mG}{\mathcal{G}\xspace}

\newcommand{\mI}{\mathcal{I}\xspace}

\newcommand{\mT}{\mathcal{T}\xspace}

\newcommand{\mX}{\mathcal{X}\xspace}

\renewcommand{\natural}{\ensuremath{\mathbb{N}}\xspace}

\usepackage{pifont}%

\renewcommand{\implies}{\rightarrow}
\newcommand{\imp}{\implies}
\newcommand{\liff}{\leftrightarrow}

\newcommand{\pattern}{{\prec\xspace}}

\newcommand\inv[1]{#1\raisebox{1.15ex}{$\scriptscriptstyle-\!1$}}

\usepackage{ stmaryrd }
\renewcommand{\imp}{\implies}
\renewcommand{\liff}{\leftrightarrow}
\newcommand{\asseq}{:=}
\newcommand{\planner}{\textsc{Patty}\xspace}
\newcommand{\N}{\mathbb{N}}
\newcommand{\Q}{\mathbb{Q}}

\newtheorem*{iexample}{Example}
\newtheorem*{rexample}{Example (cont'd)}

\newcommand{\re}{{\ensuremath{R^2\exists}}}
\newcommand{\tot}{\mathrel{<}}

\begin{document}

\maketitle

\begin{abstract}
In this paper, we propose a novel  approach for solving linear numeric planning problems, called Symbolic Pattern Planning. 
Given a planning problem $\Pi$, a bound $n$ and a pattern --defined as an arbitrary sequence of actions-- we encode the problem of finding a plan for $\Pi$ with bound $n$ as a formula with fewer variables and/or clauses than the state-of-the-art rolled-up and relaxed-relaxed-$\exists$ encodings. 
More importantly, we prove that for any given bound, it is never the case that the latter two encodings allow finding a valid plan while ours does not. 
On the experimental side, we consider 6 other planning systems  --including the ones which participated in this year's International Planning Competition (IPC)-- and we show that our planner \planner{} has remarkably good comparative performances on this year's IPC problems.
\end{abstract}

\section{Introduction}

Planning is one of the oldest problems in Artificial Intelligence, see, e.g., \cite{McCarthy1969-MCCSPP}. Starting from the classical setting in which all the variables are Boolean, in simple numeric planning problems variables can also range over the rationals and actions can increment or decrement their  values by a fixed constant, while in linear numeric  planning problems actions can also update variables to a new value which is a linear combination of the values of the variables in the state in which actions are executed, see, e.g., \cite{ipc2023}. Current approaches for solving a numeric planning problem $\Pi$ are either search-based (in which the state space is  explored using techniques based on heuristic search, see, e.g., \cite{Bonet_Geffner_2001}) or symbolic-based (in which a bound $n$ on the number of steps is a priori fixed and the problem of finding a plan with bound $n$ is encoded into a formula for which a decision procedure is available, see, e.g., \cite{DBLP:conf/ecai/KautzS92}).

In this paper, we propose a novel symbolic approach for solving numeric planning problems, called symbolic pattern planning. Given a problem $\Pi$ and a pattern $\pattern$ -- defined as a sequence of  actions -- we show how it is possible to generalize the state-of-the-art rolled-up encoding $\Pi^R$ proposed in \cite{Scala_Ramirez_Haslum_Thiebaux_2016_Rolling} and the relaxed-relaxed-$\exists$ ($\re{}$) encoding $\Pi^\re{}$ proposed in \cite{DBLP:conf/ijcai/BofillEV17}, and define a new encoding $\Pi^\pattern$  which provably dominates both $\Pi^R$ and $\Pi^\re{}$: for any bound $n$, it is never the case that the latter two allow to find a valid plan for $\Pi$ while ours does not. Further, our encoding produces formulas with fewer  clauses than the rolled-up encoding and  also with far fewer variables than the \re{} encoding, even when considering a fixed bound. Most importantly, we believe that our proposal provides a new starting point for symbolic approaches: a pattern $\pattern$ can be {\em any} sequence of actions (even with repetitions) and, assuming $n=1$, the formula produced by $\Pi^\pattern$ encodes all the sequences of actions in which each action in $\pattern$ is sequentially executed \change{0, 1}{zero, one} or possibly even more than one time. Thus, any planning problem can be solved with bound $n=1$ when considering a suitable pattern, %
and such pattern can be symbolically searched and incrementally  defined also while increasing the bound, bridging the gap between symbolic and search-based planning.

To show the effectiveness of our proposal, we $(i)$ considered the 2 planners, benchmarks, and settings of the just concluded IPC, Agile track \cite{ipc2023}; and $(ii)$ added 4 other planning systems for both simple and linear numeric problems. Overall, our comparative analysis included 6 other planners, 3 of which symbolic and 3 search-based. The results show that, compared to the other symbolic planners, our planner \planner has always better performance on every domain, 
while compared to all the other planners, \planner has overall remarkably good performances, 
being the fastest system able to solve most problems on
the largest number of domains.

The paper is structured as follows. After the preliminaries, we present the rolled-up, {\re} and our pattern encodings, and prove that the latter dominates the previous two. Then, the experimental analysis and the conclusion follow. One running example is used throughout the paper to illustrate the formal definitions and the theoretical results.

\section{Preliminaries}

We consider a fragment of numeric planning expressible
with PDDL2.1, level 2 \cite{Fox_Long_2003}. 
A \textsl{numeric planning problem} is a tuple $\Pi = \langle V_B, V_N,A, I,G\rangle$,
where $V_B$ and $V_N$ are finite sets of {\sl Boolean} and {\sl numeric
variables} with domains $\{\top, \bot\}$ and $\Q$, respectively ($\top$ and $\bot$ are the symbols we use for truth and falsity). $I$ is the {\sl initial state} mapping each variable to an element in its domain. 
A {\sl propositional condition} for a variable $v \in V_B$ is either $v = \top$ or $v=\bot$, while a {\sl numeric condition} has the
form $\psi\unrhd 0$, where $\unrhd \in \{\geq,>,=\}$ and $\psi$
is a {\sl linear expression} over $V_N$, i.e., is equal to  $\sum_{w\in V_N} k_w w + k$, for some $k_w, k
\in \Q$. 
$G$ is a finite set of {\sl goal formulas}, each one being a propositional combination of propositional and numeric conditions.
Finally, $A$
is a finite set of actions. 
An {\sl action} $a$ is a pair $\langle \op{pre}(a), \op{eff}(a)\rangle$ %
in which $(i)$ $\op{pre}(a)$ is the union of the sets of {\sl propositional} and {\sl numeric preconditions} of $a$, represented as propositional and numeric conditions, respectively; and $(ii)$ $\op{eff}(a)$ is the union of the sets of {\sl propositional} and {\sl numeric effects}, the former of the form $v \asseq \top$ or $v \asseq \bot$, the latter of the form $w \asseq \psi$, with $v \in V_B$, $w \in V_N$ and $\psi$ a linear expression. 
We assume that for each action $a$ and variable $v \in V_B\cup V_N$, $v$ occurs in $\op{eff}(a)$ at most once to the left of the operator ``$\asseq$'', and when this happens we say that $v$ is {\sl assigned} by $a$. In the rest of the paper, $v, w, x, y$  denote variables, $a, b$  denote actions and $\psi$ denotes a linear expression, each symbol possibly decorated with subscripts.

\begin{iexample}\label{ex:pddl}
    There are two robots $l$ and $r$ for left and right, respectively, whose position $x_l$ and $x_r$ on an axis correspond to the integers $\leq 0$ and $\geq 0$, respectively. The two robots can move to the left or to the right, decreasing or increasing their position by 1. The two robots carry $q_l$ and $q_r$ objects, which they can exchange. However, before exchanging objects at rate $q$, the two robots must connect setting a Boolean variable $p$ to $\top$, and this is possible only if they have the same position. Once connected, they must disconnect before moving again. The quantity $q$ can be positive or negative, corresponding to $l$ giving objects to $r$ or vice versa. This scenario can be modelled in PDDL with $V_B = \{p\}$, $V_N = \{x_l, x_r, q_l, q_r, q\}$ and the following set of actions:
{\small
\setlength{\arraycolsep}{0pt}
\begin{equation}\label{eq:ex}
\begin{array}c
    \ttt{lft}_r: \tuple{\set{x_{r} > 0}, \set{x_{r} \minuseq 1}}, \ 
    \ttt{rgt}_r : \tuple{\set{p = \bot}, \set{x_{r} \pluseq 1}},\\
    \ttt{lft}_l: \tuple{\set{p = \bot}, \set{x_{l} \minuseq 1}}, \ 
    \ttt{rgt}_l : \tuple{\set{x_{l} < 0}, \set{x_{l} \pluseq 1}},\\
        \ttt{conn}:\langle\set{x_l = x_r},
    \set{p\asseq \top} \rangle, \
        \ttt{disc}: \langle\set{p = \top},\set{p \asseq \bot} \rangle, \\
        \ttt{exch}:\langle \set{p = \top, q_{l} \geq q, q_{r} \geq -q}, 
    \set{q_{l} \minuseq q, q_{r} \pluseq q } \rangle, \\
        \ttt{lre}:\langle \set{}, \set{q \asseq 1} \rangle,
        \ttt{rle}:\langle \set{}, \set{q \asseq -1} \rangle.
\end{array}
\end{equation}
}
As customary, $v \pluseq \psi$ is an abbreviation for $v \asseq v + \psi$ and similarly for $v \minuseq \psi$ \change{}{and we abbreviate $-\psi > 0$ with the equivalent $\psi < 0$}. 
\end{iexample}

Let $\Pi= \langle V_B, V_N,A, I,G\rangle$ be a numeric planning problem. A \textsl{state} $s$ maps each variable $v\in V_B \cup V_N$ to a value $s(v)$
in its domain, and is extended to linear expressions, Boolean and numeric conditions and their propositional combinations. 
An action $a \in A$ is \textsl{executable} in a state $s$ if $s$ satisfies all the preconditions of $a$. Given a state $s$ and an executable action $a$, the \textsl{result of executing $a$ in $s$} is the state $s'$  such that for each variable $v \in V_B \cup V_N$,
 \begin{enumerate}
     \item $s'(v) = \top$ if $v \asseq \top\in \op{eff}(a)$, $s'(v) = \bot$ if $v \asseq \bot \in
\op{eff}(a)$, $s'(v) = s(\psi)$ if $(v \asseq \psi) \in \op{eff}(a)$, and
\item
$s'(v) = s(v)$ otherwise.
 \end{enumerate}

Given a finite sequence $\alpha$ of actions $a_0;\ldots;a_{n-1}$ of length $n \geq 0$, the
\textsl{state sequence $s_0; \ldots; s_n$ induced by $\alpha$ in $s_0$} is such that for $i \in [0,n)$, $s_{i+1}$ 
    $(i)$ is undefined if either $a_{i}$ is not executable in $s_i$ or $s_i$ is undefined, and 
    $(ii)$ is the result of executing $a_{i}$ in $s_i$ otherwise.

Consider a finite sequence of actions $\alpha$. We say that $\alpha$ is \textsl{executable in a state $s_0$} if each state in the sequence induced by $\alpha$  in $s_0$ is defined.  %
If $\alpha$ is executable in the initial state $I$ and the last state induced by $\alpha$ in $I$ satisfies the goal formulas in $G$, we say that $\alpha$ is a {\sl (valid) plan}. 
In the following, we will use $\alpha$ and $\pi$ to,  respectively, denote a generic sequence of actions and a plan, possibly decorated with subscripts.
For an action $a$ and $k \in \N$, $a^k$ denotes the sequence consisting of the action $a$ repeated $k$ times.

\begin{rexample}
Assume the initial state is $I = \{p = \bot, x_l = -X_I, x_r = X_I, q_l = Q, q_r = 0, q = 1\}$, where $X_I, Q$ are positive integers. Assuming $G= \set{q_l = 0, q_r = Q, x_l = -X_I, x_r = X_I}$, one of the shortest plans is 
\begin{equation}\label{eq:ex:shortest-plan}
\!\!\ttt{rgt}_l^{X_I};\ttt{lft}_r^{X_I};\ttt{conn};\ttt{exch}^{Q}; \ttt{disc}; \ttt{lft}_l^{X_I};\ttt{rgt}_r^{X_I}\!
\end{equation} 
corresponding to the robots going to the origin, connect, exchange the $Q$ items, disconnect, and then go back to their initial positions. 
\end{rexample}

\section{Symbolic Planning With Patterns}

\subsection{Symbolic Planning}

Let $\Pi = \langle V_B, V_N,A, I,G\rangle$ be a numeric planning problem.

An \textsl{encoding $\Pi^E$ of $\Pi$}
is a 5-tuple $\Pi^E = \langle \mX, \mA, \mI(\mX), \mT(\mX,\mA,\mX'), \mG(\mX)\rangle$ where
\begin{enumerate}
    \item $\mX$ is a finite set of \textsl{state variables}, each one equipped with a domain representing the values it can take. 
We assume $V_B \cup V_N \subseteq \mX$.
\item $\mA$ is a finite set of \textsl{ action variables}, each one equipped with a domain representing the values it can take.
    \item $\mI(\mX)$ is the \textsl{initial state formula}, a formula 
    in the set $\mX$ of variables defined as%
    $$
    \bigwedge_{v: I(v) = \bot} \neg v  \wedge \bigwedge_{w: I(w) = \top} w  \wedge \bigwedge_{x, k: I(x) = k} x = k.
    $$
\item
 $\mT(\mX,\mA,\mX')$ is the \textsl{symbolic transition relation}, a formula in the variables $\mX \cup \mA \cup \mX'$, where $\mX'$ is a copy of $\mX$. Together with $\mT(\mX,\mA,\mX')$, a {\sl decoding function} has to be defined enabling to associate to each model of $\mT(\mX,\mA,\mX')$ at least one sequence of actions in $A$.
 Standard requirements for $\mT(\mX,\mA,\mX')$ are:
 \begin{enumerate}
     \item {\sl correctness}: for each sequence of actions $\alpha$ corresponding to a model $\mu$ of $\mT(\mX,\mA,\mX')$, $(i)$  $\alpha$ is executable in the state $s$ \change{such that}{in which}, for each variable $v \in V_B \cup V_N$, $s(v) = \mu(v)$; and $(ii)$ the last state induced by $\alpha$ executed in $s$ is the state $s'$ such that, for each variable $v \in V_B \cup V_N$, $s'(v) = \mu(v')$;
     \item {\sl completeness}: for each state $s$ and action $a \in A$ executable in $s$ with resulting state $s'$, there must be a model $\mu$ of $\mT(\mX,\mA,\mX')$ such that, for each state variable $v \in V_B \cup V_N$, $\mu(v) = s(v)$, $\mu(v') = s'(v)$ and \change{$a$ is a sequence of actions corresponding}{the sequence of actions containing only $a$ corresponds} to $\mu$.
 \end{enumerate} 
\item $\mG(\mX)$ is the \textsl{goal formula}, 
    obtained by making the conjunction of the formulas in $G$, once $v = \top$ and $v = \bot$ are substituted with $v$ and $\neg v$, respectively.
\end{enumerate}

\begin{rexample} The initial state and goal formulas are
    $(\neg p \wedge x_l = -X_I \wedge x_r = X_I \wedge q_l = Q \wedge q_r = 0 \wedge q = 1)$ and
    $(q_l = 0 \wedge q_r = Q \wedge x_l = -X_I \wedge x_r = X_I)$, respectively.
\end{rexample}

Let $\Pi^E = \langle \mX, \mA, \mI(\mX), \mT(\mX,\mA,\mX'), \mG(\mX)\rangle$ be an encoding of $\Pi$.
As in the planning as satisfiability approach \cite{DBLP:conf/ecai/KautzS92}, we fix an integer $n \geq 0$ called {\sl bound} or {\sl number of steps}, we make $n+1$ disjoint copies  $\mX_0,\ldots,\mX_n$ of the set  $\mX$ of state variables, and $n$ disjoint copies $\mA_0,\ldots,\mA_{n-1}$ of the set $\mA$ of action variables, and define
\begin{enumerate}
    \item $\mI(\mX_0)$ as the formula in the variables $\mX_0$ obtained by substituting each  variable $x \in \mX$ with $x_0 \in \mX_0$ in $\mI(\mX)$;
    \item for each step $i = 0,\ldots,n-1$, $\mT(\mX_i,\mA_i,\mX_{i+1})$ as the formula in the variables $\mX_i \cup \mA_i \cup \mX_{i+1}$ obtained by substituting each  variable $x \in \mX$ (resp. $a \in \mA$, $x' \in \mX'$) with $x_i \in \mX_i$ (resp. $a_i \in \mA_i$, $x_{i+1} \in \mX_{i+1}$) in $\mT(\mX,\mA,\mX')$;
   \item $\mG(\mX_n)$ as the formula in the variables $\mX_n$ obtained by substituting each  variable $x \in \mX$ with $x_n \in \mX_n$ in $\mG(\mX)$.
\end{enumerate}
 Then, the \textsl{encoding $\Pi^E$ of $\Pi$ with bound $n$} is the formula 
\begin{equation}\label{eq:enc-sat}
     \Pi^E_n = \mI(\mX_0) \wedge \bigwedge_{i=0}^{n-1} \mT(\mX_i,\mA_i,\mX_{i+1}) \wedge \mG(\mX_n).
\end{equation}
To each model $\mu$ of $\Pi^E_n$, we associate the set of sequences of actions $\alpha_0; \ldots ; \alpha_{n-1}$, where each $\alpha_i$ is a sequence of actions  corresponding to the model of 
$\mT(\mX_i,\mA_i,\mX_{i+1})$ obtained by restricting $\mu$ to $\mX_i\cup\mA_i\cup\mX_{i+1}$, $i \in [0,n)$. In the following,  \change{$\Pi^{E -1}_n$}{$\inv{(\Pi^E_n)}$} is  the set of sequences of actions in $A$ associated to a model of $\Pi^E_n$. 
The correctness of $\mT(\mX,\mA,\mX')$ ensures the {\sl correctness of $\Pi^E$}: for each bound $n$, each sequence in \change{$\Pi^{E-1}_n$}{$\inv{(\Pi^E_n)}$} is a plan. 
The completeness of $\mT(\mX,\mA,\mX')$ ensures the {\sl completeness of $\Pi^E$}: if it exists a plan for $\Pi$, it will be found by considering $\Pi^E_0$, $\Pi^E_1$, \dots. 

It is clear that the number of variables and size of (\ref{eq:enc-sat}) increase with the bound $n$, explaining why much of the research has concentrated on how to produce encodings allowing to find plans with the lowest possible bound $n$.

\subsection{Rolled-up, Standard  and {\re} Encodings}

Let $\Pi = \langle V_B, V_N,A, I,G\rangle$ be a numeric planning problem. Many encodings have been proposed, each characterized by how the symbolic transition relation 
is computed. In most encodings (see, e.g., \cite{DBLP:journals/ai/RintanenHN06,DBLP:conf/ijcai/BofillEV17,Leofante2020}), each action $a \in A$ is defined as a Boolean variable in $\mA$ which will be true (resp. false) in a model $\mu$ of $\mT(\mX,\mA,\mX')$ if action $a$ occurs once (resp. does not occur) in each sequence of actions corresponding to $\mu$. 
Here we start presenting the state-of-the-art {\sl rolled-up encoding $\Pi^R$ of $\Pi$} proposed by \cite{Scala_Ramirez_Haslum_Thiebaux_2016_Rolling}. In $\Pi^R$, each action $a \in A$ is defined as an action variable which can get an arbitrary value $k \in \N$, and this corresponds to have $k$ (consecutive) occurrences of $a$ in the action sequences corresponding to the models of the symbolic transition relation of $\Pi^R$.
\footnote{To ease the presentation,  our definition of  $\Pi^R$ considers just the cases $\alpha=0$ and $\alpha=1$ of Theorem~1 in \cite{Scala_Ramirez_Haslum_Thiebaux_2016_Rolling}, which (quoting) ‘‘cover a very general class of dynamics, where rates
of change are described by linear or constant equations".} 
However, in $\Pi^R$ it is not the case that each action $a$ can get a value $>1$, (e.g., because $a$ cannot be executed more than once, or it is not useful to execute $a$ more than once), and the definition of when it is possible to set  $a > 1$ depends on the form of the effects of $a$. For this reason, each effect $v \asseq e$ of an action  $a$ is categorized as
\begin{enumerate}
    \item a {\sl Boolean assignment}, if $v \in V_B$ and $e \in \{\top,\bot\}$, as for the effects of the actions \ttt{conn} and \ttt{disc} in (\ref{eq:ex}), or as
    \item a {\sl linear increment}, if $e = v + \psi$ with $\psi$ a linear expression not containing any of the variables assigned by $a$,
    as for the effects of the action
$\ttt{exch}$ and 
$\ttt{lft}_r$ in (\ref{eq:ex}), or as
    \item a {\sl general assignment}, if it does not fall in the above two categories. General assignments are further divided into
    \begin{enumerate}
        \item {\sl simple assignments}, when $e$ does not contain any of the variables assigned by $a$, as in the effects of the actions $\ttt{lre}$ and $\ttt{rle}$ in (\ref{eq:ex}), and
        \item {\sl self-interfering assignments}, otherwise.
    \end{enumerate}
\end{enumerate}
Then, an action $a$ is {\sl eligible for rolling} if 
\begin{enumerate}
    \item 
    $v = \bot\in \op{pre}(a)$ (resp. $v = \top\in \op{pre}(a)$) implies $v \asseq \top\not\in \op{eff}(a)$ (resp. $v \asseq \bot\not\in \op{eff}(a)$), and
    \item $a$ does not contain a self-interfering assignment, and
    \item $a$ contains a linear increment.
\end{enumerate}
The result of rolling action $a$ for $k\geq 1$ times is such that
\begin{enumerate}
    \item if $v \pluseq \psi \in \ttt{eff}(a)$ is a linear increment, then the value of $v$ is incremented by $k \times \psi$, while
    \item if $v \asseq e \in \ttt{eff}(a)$ is a Boolean or simple assignment, then the value of $v$ becomes $e$, equal to the value obtained after a single execution of $a$.
\end{enumerate}
On the other hand, if an action $a$ is not eligible for rolling,  $a > 1$ is not allowed, and this can be enforced through at-most-once (“amo”) axioms.

In $\Pi^R$, the symbolic transition relation $\mT^R(\mX,\mA,\mX')$ is the conjunction of the formulas in the following sets: 
\begin{enumerate}
    \item $\op{pre}^R(A)$, consisting of, for each $a \in A$, $v = \bot$ and $w = \top$ in $\op{pre}(a)$,
    $$
    a > 0 \implies (\neg v \wedge w),%
   \footnote{We do not use the equivalent formulation $(a > 0 \implies \neg v)$, $(a > 0 \implies w)$, which has a more direct translation to clauses, in order to save space. Analogously in the rest of the paper.}
    $$
    and, for each $a \in A$ and $\psi \unrhd 0$ in $\op{pre}(a)$,
    $$
    a > 0 \imp \psi \unrhd 0, \qquad a > 1 \imp \psi[a] \unrhd 0,
    $$
    where $\psi[a]$ is the linear expression obtained from $\psi$ by substituting
    each variable $x$ with 
    \begin{enumerate}
        \item $x + (a-1) \times \psi_1$, whenever $x \pluseq \psi_1 \in \ttt{eff}(a)$ is a linear increment,
        \item $\psi_1$, if $x \asseq \psi_1 \in \op{eff}(a)$ is a simple assignment.
    \end{enumerate}
    The  last two formulas ensure that $\psi \unrhd 0$  holds in the states in which the first and the last execution of $a$ happens (see \cite{Scala_Ramirez_Haslum_Thiebaux_2016_Rolling}). 
    
    \item $\op{eff}^R(A)$, consisting of, for each $a \in A$, $v \asseq \bot$, $w \asseq \top$, linear increment $x \pluseq  \psi$ and
    general assignment $y  \asseq \psi_1$ 
    in $\op{eff}(a)$,
    $$
    a > 0 \implies  (\neg v' \wedge w' \wedge x' = x + a \times \psi \wedge y' = \psi_1).
    $$
    \item $\op{frame}^R(V_B\cup V_N)$, consisting of, for each variable $v \in V_B$ and $w \in V_N$, 
    $$
    \begin{array}{c}
   \!\!\!(\bigwedge_{a: v \asseq \top \in \op{eff}(a)} a = 0 \wedge
    \bigwedge_{a: v \asseq \bot \in \op{eff}(a)} a = 0) \imp v' \equiv v, \\
    \bigwedge_{a : w \asseq \psi \in \op{eff}(a)} a = 0 \imp w' = w.
    \end{array}
    $$
    \item $\op{mutex}^R(A)$ consisting of 
    $(a_1 = 0 \vee a_2 = 0),$
    for each pair of distinct actions $a_1$ and $a_2$ such that there exists a variable $v$ with
    \begin{enumerate}
        \item $v \in V_B$, $v = \bot$ (resp. $v = \top$) in $\op{pre}(a_1)$ and $v \asseq \top$ (resp. $v \asseq \bot$) in $\op{eff}(a_2)$, or 
        \item $v \in V_N$, $v \asseq \psi \in \ttt{eff}(a_1)$ and $v$ occurring either in $\ttt{eff}(a_2)$ or in $\ttt{pre}(a_2)$.
    \end{enumerate}
    \item $\op{amo}^R(A)$ consisting of, for each action $a$ not eligible for rolling, 
    $$(a = 0 \vee a = 1).$$
\end{enumerate}
Notice that if for action $a$ the formula $(a=0 \vee a=1)$ belongs to $\mT^R(\mX,\mA,\mX')$, we can equivalently
$(i)$ define $a$ to be a Boolean variable, and then
$(ii)$ replace $a=0$, $a > 0$, $a=1$ and $a > 1$ with $\neg a$, $a$, $a$ and $\bot$, respectively, in $\mT^R(\mX,\mA,\mX')$.
It is clear that if $\mT^R(\mX,\mA,\mX')$ contains $(a = 0 \vee a = 1)$ for any action $a$, then the rolled-up encoding $\Pi^R$ reduces to the standard encoding as defined, e.g., in \cite{Leofante2020}. Equivalently,  in the {\sl standard encoding $\Pi^S$ of $\Pi$}, the symbolic transition relation $\mT^S(\mX,\mA,\mX')$ is obtained by adding, for each action $a$, $(a = 0 \vee a = 1)$ to $\mT^R(\mX,\mA,\mX')$.
The decoding function of the rolled-up (resp. standard) encoding associates to each model $\mu$ of $\mT^R(\mX,\mA,\mX')$ (resp. $\mT^S(\mX,\mA,\mX')$) the sequences of actions in which each action $a$ occurs $\mu(a)$ times. 

The biggest problem with the rolled-up and standard encodings is the presence of the axioms in $\op{mutex}(A)$, which $(i)$ cause the size of $\mT^R(\mX,\mA,\mX')$ to be possibly quadratic in the size of $\Pi$; and $(ii)$ forces some actions to be set to $0$ even when it is not necessary to maintain the correctness and completeness of $\mT^R(\mX,\mA,\mX')$, see, e.g., \cite{DBLP:journals/ai/RintanenHN06}. 
Indeed, allowing to set more actions to a value $> 0$ while maintaining correctness and completeness, allows finding solutions to (\ref{eq:enc-sat}) with a lower value for the bound.
Several proposals along these lines have been made. Here we present the \re{} 
encoding presented in \cite{DBLP:conf/ijcai/BofillEV17} which is arguably the state-of-the-art encoding  in which actions are encoded as Boolean variables (though there \change{exists}{exist} cases in which the $\exists$-encoding presented in \cite{DBLP:journals/ai/RintanenHN06} allows to solve (\ref{eq:enc-sat}) with a value for the bound lower than the one needed by the \re{} encoding).

In the \re{} encoding, action variables are Boolean and assumed to be  ordered according to a given total order $\tot$. In general, different orderings lead to different \re{} encodings. In the following, we represent and reason about $\tot$ considering the corresponding sequence of actions (which indeed contains each action in $A$ exactly once) and define $\Pi^<$ to be the {\sl \re{} $<$-encoding of $\Pi$}. In $\Pi^<$, for each action $a$ and variable $v$ assigned by $a$, a newly introduced variable $v^a$ with the same domain of $v$  is added to the set $\mX$ of state variables.  Intuitively, each new variable $v^a$
represents the value of $v$ after the sequential execution of some actions in the initial sequence of $<$ ending with $a$.  
The symbolic transition relation $\mT^<(\mX,\mA,\mX')$ of $\Pi^<$ is the conjunction of the formulas in the following sets: 
\begin{enumerate}
    \item $\op{pre}^<(A)$, consisting of, for each $a \in A$, $v = \bot$, $w = \top$ and $\psi \unrhd 0$ in $\op{pre}(a)$,
    $$
    a \imp (\neg v^{\ll,a} \wedge w ^{\ll,a} \wedge \psi^{\ll,a} \unrhd 0),
    $$
    where, for each variable $x \in V_B \cup V_N$, $x^{\ll,a}$  stands for the variable $(i)$ $x$, if there is no action preceding $a$ in $<$ assigning $x$; and $(ii)$
    $x^b$, if $b$ is the last action assigning $x$ preceding $a$ in $<$. Analogously, $\psi^{\ll,a}$ is the linear expression obtained from $\psi$ by substituting
    each variable $x \in V_N$ with $x^{\ll,a}$.
    \item $\op{eff}^<(A)$, consisting of, for each $a \in A$, $v \asseq \bot$, $w \asseq \top$ and general assignment $x  \asseq \psi$ in $\op{eff}(a)$,
    $$
    \begin{array}{cc}
    a \implies  (\neg v^a \wedge w^a \wedge x^a = \psi^{\ll,a}), \\
    \neg a \implies (v^a \liff v^{\ll,a} \wedge 
     w^a \liff w^{\ll,a} \wedge 
     x^a = x^{\ll,a}).
   \end{array}
   $$
    \item $\op{frame}^<(V_B\cup V_N)$, consisting of, for each variable $v \in V_B$ and $w \in V_N$, 
    $$
    v' \liff v^{\ll,g}, \qquad w' = w^{\ll,g},
    $$
    where $g$ is a dummy action assumed to follow all the other actions in $<$.
\end{enumerate}
The decoding function of the \re{} $<$-encoding associates to each model $\mu$ of $\mT^<(\mX,\mA,\mX')$ the sequence of actions obtained from $<$ by deleting the actions $a$ with $\mu(a)=\bot$. 
In the \re{} $<$-encoding, there are no mutex axioms and  the size of $\mT^<(\mX,\mA,\mX')$ is linear in the size of  $\Pi$. However, as we mentioned previously, it introduces many new state variables (in the worst case, $|V_B \cup V_N| \times |A|$).

The main advantage of $\Pi^R$ and $\Pi^<$ over $\Pi^S$ is that the first two allow to find plans with lower values for the bound.

\begin{rexample}
    The rolled-up (resp. standard) encoding of the two robots problem admits a model with bound $n=5$ (resp. $n=2X_I+Q+2$, and thus $n=5$ when $X_I = Q = 1$). The \re{} $<$-encoding admits a model with bound $n=2(X_I-1)+Q$ if actions in $<$ are ordered as in the plan (\ref{eq:ex:shortest-plan}), and thus $n=1$ when $X_I = Q = 1$. In the worst case, the \re{} $<$-encoding admits a solution with a bound equal to the one needed by the standard encoding, and this happens when actions in $<$ are in reverse order wrt the plan~(\ref{eq:ex:shortest-plan}). 
\end{rexample}
As the example shows, $\Pi^{R}$ and $\Pi^<$ dominate $\Pi^{S}$, while $\Pi^{R}$ and $\Pi^<$ do not dominate each other.
Given two correct encodings $\Pi^{E_1}$ and $\Pi^{E_2}$ of $\Pi$, $\Pi^{E_1}$ {\sl dominates} $\Pi^{E_2}$ if for any bound $n$, $\Pi^{E_2}_n$ satisfiability implies $\Pi^{E_1}_n$ satisfiability. 

\begin{theorem}
    Let $\Pi$ be a numeric planning problem. Let $<$ be a total order of actions. The rolled-up  encoding $\Pi^R$, the \re{} $<$-encoding $\Pi^<$ and the standard encoding $\Pi^S$ of $\Pi$ are correct and complete.
    $\Pi^R$ and $\Pi^<$ dominate $\Pi^S$.
\end{theorem}

\begin{proof}(Sketch)
For the correctness of $\Pi^R$ (and thus of $\Pi^S$) and $\Pi^<$ see Prop. 3 and Theorem 1 in the respective original papers. The completeness of $\Pi^S$ is taken for granted. A model of $\Pi_n^S$ with a corresponding plan $\pi$ is also a model of $\Pi_n^R$, and can be easily used to define a model of $\Pi_n^<$ with the same corresponding plan $\pi$. This
implies the completeness of $\Pi^R$ and $\Pi^<$ and the fact that they dominate $\Pi^S$. 
\end{proof}

\subsection{Pattern Encoding}

Let $\Pi = \langle V_B, V_N,A, I,G\rangle$ be a numeric planning problem.
In the pattern encoding we combine and then generalize the strengths of the rolled-up and \re{} encoding by $(i)$ allowing for the multiple executions of actions; $(ii)$ considering an ordering to avoid mutexes; and $(iii)$  allowing for {\em arbitrary} sequences of actions.

Consider a {\sl pattern $\pattern$}, defined as a possibly empty, finite sequence of actions. 
In the {\sl pattern $\pattern$-encoding $\Pi^\pattern$ of $\Pi$},
\begin{enumerate}
    \item $\mX = V_B \cup V_N \cup V$, where $V$ contains a newly introduced variable $v^{\pattern_1;a}$ with the same range of $v$, for each variable $v$ and initial pattern $\pattern_1;a$ of $\pattern$ \change{}{(i.e., $\pattern$ starts with $\pattern_1;a$)} in which $a$ contains a general assignment of $v$, and
    \item $\mA$ contains one action variable $a^{\pattern_1}$ ranging over $\N$, for each initial pattern $\pattern_1;a$ of $\pattern$.
\end{enumerate}
Then, the value of a variable $v\in V_B\cup V_N$ after one or more  of the actions in $\pattern$ are executed  (possibly consecutively multiple times) is given by $\sigma^\pattern(v)$, where $\sigma^\pattern(v)$ is inductively defined as
$(i)$ $\sigma^{\pattern}(v) = v$ if $\pattern$ is the empty sequence; and $(ii)$ for a non-empty pattern $\pattern = \pattern_1;a$,
\begin{enumerate}
\item 
if $v$ is not assigned by $a$,  $\sigma^{\pattern}(v) = \sigma^{\pattern_1}(v)$;
    \item 
if $v \asseq \top \in 
      \op{eff}(a)$, $\sigma^{\pattern}(v) = (\sigma^{\pattern_1}(v) \vee a > 0)$;
   \item if $v \asseq \bot \in 
      \op{eff}(a)$, $\sigma^{\pattern}(v) = (\sigma^{\pattern_1}(v) \wedge a = 0)$;
    \item 
if $v \pluseq \psi \in \op{eff}(a)$ is a linear increment, 
        $\sigma^{\pattern}(v) = \sigma^{\pattern_1}(v) + a \times \sigma^{\pattern_1}(\psi)$;
        \item 
        if $v \asseq \psi \in \op{eff}(a)$ is a general assignment
        $\sigma^{\pattern}(v) = v^{\pattern}$.
\end{enumerate}
Above and in the following, for any pattern $\pattern_1$ and linear expression $\psi$, $\sigma^{\pattern_1}(\psi)$ is the expression obtained by substituting each variable $v \in V_N$ in $\psi$ with $\sigma^{\pattern_1}(v)$.

\begin{rexample}
Consider (\ref{eq:ex}), 
and assume $\pattern$ is
$$
  \ttt{lre};\ttt{rle};\ttt{lft}_r;\ttt{rgt}_l;\ttt{conn};\ttt{exch}; \ttt{disc};\ttt{rgt}_r; \ttt{lft}_l.
$$
We have two newly introduced variables $q^{lre}$ and $q^{lre;rle}$, and for the Boolean variable $p$,
$$
\sigma^\pattern(p) = (p \vee \ttt{conn} > 0) \wedge \ttt{disc} = 0,
$$
and, for the
numeric variables in
$V_N = \{x_l, x_r, q_l, q_r, q\}$,
$$
\begin{array}{cc}
\sigma^\pattern(x_l) = x_l + \ttt{rgt}_l - \ttt{lft}_l, \\
\sigma^\pattern(x_r) = x_r - \ttt{lft}_r + \ttt{rgt}_r, \\
\sigma^\pattern(q_l) = q_l - \ttt{exch} \times q^{lre;rle}, \\
\sigma^\pattern(q_r) = q_r + \ttt{exch} \times q^{lre;rle}, \\
\sigma^\pattern(q) = q^{\ttt{lre;rle}}.
\end{array}
$$
\end{rexample}

The symbolic transition relation $\mT^\pattern(\mX,\mA,\mX')$ of $\Pi^\pattern$ is the conjunction of the formulas in the following sets:
\begin{enumerate}

\item $\op{pre}^\pattern(A)$, which contains, for each initial pattern $\pattern_1; a$ of $\pattern$, and for each 
 $v = \bot$ and $w = \top$ in $\op{pre}(a)$,
\begin{equation*}\begin{array}{c}
    a^{\pattern_1} > 0 \implies (\neg \sigma^{\pattern_1}(v)
    \wedge \sigma^{\pattern_1}(w)),
\end{array} \end{equation*} 
and, for each numeric precondition $\psi \unrhd 0$ in $\op{pre}(a)$, 
\begin{equation*}\begin{array}{c}
    a^{\pattern_1} > 0 \imp \sigma^{\pattern_1}(\psi) \unrhd 0, \ \
    a^{\pattern_1} > 1 \imp \sigma^{\pattern_1}(\psi[a]) \unrhd 0.
\end{array} \end{equation*}  

\item $\op{eff}^\pattern(A)$, consisting of, for each initial pattern $\pattern_1; a$ of $\pattern$ and variable $v$ such that 
$v \asseq \psi \in \op{eff}(a)$ is a general assignment,
    \begin{equation*}\begin{array}{c}
        a^{\pattern_1} = 0 \implies v^{\pattern_1;a} = \sigma^{\pattern_1}(v), \\
        a^{\pattern_1} > 0 \implies v^{\pattern_1;a} = \sigma^{\pattern_1}(\psi).
    \end{array} \end{equation*}  

    \item 
    $\op{amo}^\pattern(A)$ which contains, for each initial pattern $\pattern_1; a$ of $\pattern$ in which $a$ is not eligible for rolling,
    \begin{equation*}
        a^{\pattern_1} = 0 \vee a^{\pattern_1}=1.
    \end{equation*}    
\item $\op{frame}^\pattern(V_B\cup V_N)$, consisting of, for each variable $v \in V_B$ and $w \in V_N$, 
\begin{equation*}\begin{array}{c}
v' \liff \sigma^\pattern(v), \qquad w' = \sigma^\pattern(w).
\end{array} \end{equation*}     
\end{enumerate}
If $\pattern = a_1; a_2; \ldots; a_k$ (with $a_1, a_2, \ldots a_k \in A$, $k \geq 0$), 
the decoding function associates to each model $\mu$ of $\mT^\pattern(\mX,\mA,\mX')$ the sequence of actions $a_1^{\mu(a_1)}; a_2^{\mu(a_2)}; \ldots; a_k^{\mu(a_k)}$, i.e., the sequence of actions listed as in $\pattern$, each action $a$ repeated $\mu(a)$ times. Notice the similarities and differences with $\Pi_n^R$ and $\Pi^<_n$. In particular, our encoding $(i)$ does not include the mutex axioms; and $(ii)$ introduces variables only when there are general assignments (usually very few, though in the worst case, $|V_B \cup V_N| \times |A|$).

Notice also that  we did not make any assumption about the pattern $\pattern$, which can be any arbitrary sequence of actions. In particular, $\pattern$ can contain multiple non-consecutive occurrences of any action $a$: this allows for models of $\mT^\pattern(\mX,\mA,\mX')$ corresponding to sequences of actions in which $a$ has multiple non-consecutive occurrences. At the same time, $\pattern$ may also not include some action $a \in A$: in this case, our encoding is not complete (unless $a$ is never executable). Even further, it is possible to consider multiple different patterns $\pattern_1,\ldots,\pattern_n$, each leading to a corresponding symbolic transition relation $\mT^{\pattern_i}(\mX,\mA,\mX')$, and then consider the encoding  (\ref{eq:enc-sat}) with bound $n$ in which $\mT(\mX_i,\mA_i,\mX_{i+1})$ is replaced by $\mT^{\pattern_i}(\mX_i,\mA_i,\mX_{i+1})$: in this case each model of the resulting encoding with bound $n$ will still correspond to a valid plan, (though we may fail to find plans with $n$ or fewer actions).
Even more, with a suitable pattern\change{$\pattern$-encodings}{$\pattern$}, any planning problem can be solved with bound $n = 1$ \change{and such}{. Such} pattern $\pattern$ can be symbolically searched and incrementally defined \change{also}{} while increasing the bound, bridging the gap between symbolic and search-based planning. Such outlined opportunities significantly extend the possibilities offered by all the other encodings, and for this reason, we believe our proposal  provides a new starting point for the research in symbolic planning. 

Here we focus on $\pattern$-encodings with bound $n$ in which we have a single, a priori fixed, simple and complete pattern. A pattern is {\sl simple} if each action occurs at most once, and is {\sl complete} if each action occurs at least once. 
If $\pattern_1;a$ is a simple pattern, $a^{\pattern_1} \in \mA$ 
(resp. $v^{\pattern_1;a} \in V$) can be abbreviated to $a$ (resp. $v^a$) without introducing ambiguities, as we do in the example below.

\begin{rexample}
In our case,  the given pattern $\pattern$ is simple and also complete, and 
$\op{pre}^\pattern(A)$ is equivalent to
\begin{equation*}\begin{array}{c}
    \ttt{lft}_r > 0 \imp x_r > 0, \; \ttt{lft}_r > 1 \imp x_r - (\ttt{lft}_r - 1) > 0, \\
    \ttt{rgt}_r > 0 \imp \neg ((p \vee \ttt{conn} > 0) \wedge \ttt{disc}= 0), \\
    \ttt{lft}_l > 0 \imp \neg ((p \vee \ttt{conn} > 0) \wedge \ttt{disc}= 0), \\    
    \ttt{rgt}_l > 0 \imp x_l < 0, \; \ttt{rgt}_l > 1 \imp x_l + (\ttt{rgt}_l - 1) < 0, \\
    \ttt{conn} > 0 \imp x_l + \ttt{rgt}_l = x_r - \ttt{lft}_r, \\
    \ttt{disc} > 0 \imp (p \vee \ttt{conn} > 0), \\
    \hspace*{-3mm}\ttt{exch} > 0 \imp ((p \vee \ttt{conn} > 0) \wedge q_l \geq q^{\ttt{rle}} \wedge  q_r \geq -q^{\ttt{rle}}),\\
    \ttt{exch} > 1 \imp (q_l \geq q^{\ttt{rle}} - (\ttt{exch} -1) \times q^{\ttt{rle}}), \\
    \ttt{exch} > 1 \imp (q_r \geq -q^{\ttt{rle}} + (\ttt{exch} -1) \times q^{\ttt{rle}}).
\end{array} \end{equation*}  

\noindent 
$\op{eff}^\pattern(A)$ is
\begin{equation*}\begin{array}{c}
    \ttt{lre} = 0 \imp q^{\ttt{lre}} = q, \qquad    \ttt{lre} > 0 \imp q^{\ttt{lre}} = 1, \\
    \ttt{rle} = 0 \imp q^{\ttt{rle}} = q^{\ttt{lre}}, \qquad \ttt{rle} > 0 \imp q^{\ttt{rle}} = -1.
\end{array} \end{equation*}  
$\op{amo}^\pattern(A)$ is 
\begin{equation*}\begin{array}{c}
    \ttt{lre} = 0 \vee \ttt{lre} = 1, \qquad 
    \ttt{rle} = 0 \vee \ttt{rle} = 1, \\
    \ttt{conn} = 0 \vee \ttt{conn} = 1, \qquad 
    \ttt{disc} = 0 \vee \ttt{disc} = 1.
\end{array} \end{equation*} 
$\op{frame}^\pattern(V_B\cup V_N)$ is
$$
\begin{array}{c}
    p' \liff ((p \vee \ttt{conn} > 0) \wedge \ttt{disc} = 0),\\
x'_l = x_l + \ttt{rgt}_l - \ttt{lft}_l, \quad x'_r = x_r - \ttt{lft}_r + \ttt{rgt}_r, \\
q'_l = q_l - \ttt{exch} \times q^{\ttt{rle}}, \quad  q'_r = q_r + \ttt{exch} \times q^{rle}, \\
q' = q^{\ttt{rle}}.
\end{array}
$$
The plan (\ref{eq:ex:shortest-plan}) belongs to \change{$\Pi^{\pattern -1}_1$}{$\inv{(\Pi^\pattern_1)}$}.
\end{rexample}
Indeed, in the case of the example, the chosen pattern allows finding a plan with bound $n=1$, compared to the rolled-up and standard encodings which need at least $n=5$, while any \re{} encoding needs a bound of at least $2(X_I - 1) + Q$ which is equal to 1 only if $X_I = Q = 1$.
Of course, as for the \re{} encoding, depending on the selected pattern, we get different results. However, our pattern $\pattern$-encoding dominates any \re{} $<$-encoding, of course if $<$ is compatible with $\pattern$. A total order $<$ \change{on}{of} actions is {\sl compatible} with $\pattern$ if $<$ (seen a sequence of actions) can be obtained from $\pattern$ by removing $0$ or more actions.

\begin{table*}[tb]
        \centering
        \resizebox{\textwidth}{!}{
        {\Huge
\begin{tabular}{|l|cccccc|cccccc|ccc|ccc|ccc|}
\hline
 & \multicolumn{6}{c|}{Coverage (\%)} & \multicolumn{6}{c|}{Time (s)} & \multicolumn{3}{c|}{Bound} & \multicolumn{3}{c|}{$|X \cup A \cup X'|$} & \multicolumn{3}{c|}{$|T(X,A,X')|$} \\
Domain & P & $R^2\exists$ & SR & EN & FF & NFD & P & $R^2\exists$ & SR & EN & FF & NFD & P & $R^2\exists$ & SR & P & $R^2\exists$ & SR & P & $R^2\exists$ & SR \\ \hline
{BlGroup} (S) & \textbf{100} & 65 & \textbf{100} & \textbf{100} & 10 & - & \textbf{1.5} & 126.5 & 2.1 & 48.0 & 270.2 & - & \textbf{1.0} & 6.0 & \textbf{1.0} & \textbf{40} & 250 & \textbf{40} & \textbf{101} & 331 & 122 \\
{Counters} (S) & \textbf{100} & 60 & \textbf{100} & \textbf{100} & 60 & 50 & \textbf{0.8} & 153.4 & 1.1 & 6.9 & 129.0 & 149.1 & \textbf{1.0} & 14.8 & \textbf{1.0} & \textbf{83} & 1.3k & \textbf{83} & \textbf{185} & 1.4k & 250 \\
{Counters} (L) & \textbf{95} & 60 & 35 & 45 & 40 & 25 & \textbf{4.6} & 152.2 & 204.1 & 180.5 & 180.0 & 225.4 & \textbf{2.0} & \textbf{2.0} & 2.5 & \textbf{26} & 125 & \textbf{26} & \textbf{58} & 169 & 112 \\
{Drone} (S) & 25 & 15 & 15 & \textbf{85} & 10 & 80 & 242.8 & 255.6 & 257.2 & \textbf{59.9} & 270.0 & 65.4 & \textbf{4.7} & 7.7 & 9.7 & 30 & 146 & \textbf{29} & \textbf{64} & 191 & 211 \\
{Watering} (S) & 25 & - & - & \textbf{100} & 10 & 60 & 226.8 & - & - & \textbf{9.8} & 276.5 & 185.2 & \textbf{8.4} & - & - & \textbf{61} & 540 & \textbf{61} & \textbf{145} & 654 & 610 \\
{Farmland} (S) & \textbf{100} & - & \textbf{100} & \textbf{100} & 35 & 75 & 0.9 & - & 1.6 & \textbf{0.7} & 206.8 & 85.5 & \textbf{1.0} & - & 2.2 & \textbf{63} & 690 & \textbf{63} & \textbf{120} & 773 & 501 \\
{Farmland} (L) & \textbf{100} & 10 & - & 75 & 75 & 55 & \textbf{1.6} & 275.1 & - & 96.8 & 90.7 & 151.7 & \textbf{1.0} & 8.0 & - & 19 & 61 & \textbf{17} & \textbf{32} & 79 & 62 \\
{HPower} (S) & \textbf{100} & 25 & - & 10 & 5 & 5 & \textbf{14.8} & 233.3 & - & 270.4 & 285.0 & 285.1 & \textbf{1.0} & \textbf{1.0} & - & \textbf{448} & 22k & \textbf{448} & \textbf{788} & 23k & 11k \\
{Sailing} (S) & \textbf{100} & - & 90 & \textbf{100} & 5 & 50 & \textbf{1.0} & - & 20.0 & 1.4 & 285.0 & 150.3 & \textbf{3.2} & - & 7.2 & \textbf{49} & 380 & \textbf{49} & \textbf{86} & 434 & 293 \\
{Sailing} (L) & \textbf{95} & 5 & - & 20 & 40 & 70 & \textbf{1.0} & 297.9 & - & 241.2 & 182.9 & 109.4 & \textbf{1.0} & 5.0 & - & 84 & 951 & \textbf{82} & \textbf{200} & 1.1k & 490 \\ \hline
{Delivery} (S) & 25 & 20 & - & 65 & \textbf{95} & 45 & 232.7 & 256.0 & - & 121.2 & \textbf{48.5} & 165.2 & \textbf{1.0} & 2.0 & - & \textbf{250} & 8.0k & - & \textbf{662} & 8.5k & - \\
{Expedit.} (S) & \textbf{15} & 5 & - & 10 & - & \textbf{15} & \textbf{253.5} & 289.0 & - & 270.3 & - & 253.7 & \textbf{5.0} & 10.0 & - & \textbf{105} & 1.5k & - & \textbf{225} & 1.6k & - \\
{MPrime} (S) & 55 & 35 & 50 & \textbf{85} & 80 & 65 & 139.7 & 205.4 & 171.2 & 49.7 & \textbf{47.5} & 133.6 & \textbf{1.5} & \textbf{1.5} & 5.2 & \textbf{467} & 39k & \textbf{467} & \textbf{1.2k} & 39k & 19k \\
{Pathways} (S) & \textbf{100} & 5 & 5 & 60 & 50 & 5 & \textbf{4.7} & 286.7 & 286.4 & 133.9 & 154.9 & 285.0 & \textbf{1.0} & 6.0 & 3.0 & \textbf{186} & 3.3k & \textbf{186} & \textbf{318} & 3.5k & 521 \\
{Rover} (S) & \textbf{85} & 45 & 55 & 35 & 50 & 20 & \textbf{77.6} & 194.5 & 185.5 & 204.4 & 142.1 & 241.0 & \textbf{1.9} & 2.0 & 7.7 & \textbf{360} & 20k & 367 & \textbf{754} & 20k & 10k \\
{Satellite} (S) & 10 & 5 & 15 & \textbf{30} & 20 & 20 & 277.3 & 292.6 & 267.7 & \textbf{222.6} & 229.4 & 242.2 & \textbf{4.0} & \textbf{4.0} & 10.0 & 222 & 7.4k & \textbf{183} & \textbf{566} & 7.8k & 5.4k \\
{Sugar} (S) & \textbf{100} & 25 & - & 95 & 65 & 25 & \textbf{6.8} & 247.2 & - & 23.7 & 119.9 & 232.9 & \textbf{2.0} & 2.2 & - & \textbf{495} & 31k & - & \textbf{1124} & 32k & - \\
{TPP} (L) & 10 & 5 & - & \textbf{20} & 10 & 10 & 275.3 & 284.4 & - & \textbf{244.3} & 268.4 & 270.0 & \textbf{3.0} & \textbf{3.0} & - & 355 & 10k & \textbf{278} & \textbf{917} & 10k & 4.0k \\
{Zeno} (S) & 55 & 55 & - & \textbf{100} & 55 & 45 & 119.2 & 129.8 & - & \textbf{20.4} & 135.0 & 178.5 & \textbf{2.1} & 2.3 & - & \textbf{198} & 6.2k & - & \textbf{577} & 6.6k & - \\ \hline
\textit{Total} & \textbf{12} & 0 & 3 & 10 & 1 & 1 & \textbf{11} & 0 & 0 & 6 & 2 & 0 & \textbf{19} & 5 & 2 & \textbf{14} & 0 & \textbf{14} & \textbf{18} & 0 & 0 \\ \hline
\end{tabular}}}
    \caption{Comparative analysis between the {Patty} (P) planner, the symbolic planners {\re{}} (\re), {SpringRoll} (SR) and the search-based planners {ENHSP} (EN), {MetricFF} (FF) and {NumericFastDownward} (NFD). The labels S and L specify if the domain presents simple or linear effects, respectively, see \cite{ipc2023}. ``k" means $\times 1000$
    }
    \vspace{-5mm}
    \label{tab:experiments}
    \end{table*}

\begin{theorem} \label{thm:domination}
Let $\Pi$ be a numeric planning problem. Let $\pattern$ be a pattern.  
\begin{enumerate}
    \item $\Pi^\pattern$ is correct.
    \item For any action $a$, $\Pi^{\pattern;a}$ dominates $\Pi^\pattern$.
    \item If $\pattern$ is complete, then $\Pi^\pattern$ is complete.
    \item If $\pattern$ is complete, then $\Pi^\pattern$ dominates   $\Pi^R$. 
    \item 
    If $<$ is a total order compatible with $\pattern$, then $\Pi^\pattern$ dominates $\Pi^<$ . 
\end{enumerate}
\end{theorem}
\begin{proof}(Sketch)
The correctness of $\Pi^\pattern$ follows from the correctness of $\mT^\pattern(\mX,\mA,\mX')$ which can be proved by induction on the length $k$ of $\pattern$:
if $k=0$ is trivial, if $k > 0$ the thesis follows from the induction hypothesis, mimicking the proof of Proposition 3 in \cite{Scala_Ramirez_Haslum_Thiebaux_2016_Rolling}.

$\Pi^{\pattern;a}$ dominates $\Pi^\pattern$, since each model $\mu$ of $\mT^{\pattern}$ can be extended to a model $\mu'$ of $\mT^{\pattern;a}$ with $\mu'(a^\pattern) = 0$.

If $\pattern$ is complete,  $\Pi^\pattern$  completeness follows from its correctness, the completeness of $\Pi^R$, and $\Pi^\pattern$ dominates $\Pi^R$.

$\Pi^\pattern$ dominates $\Pi^R$ because for each model $\mu^R$ of $\mT^R$ we can define a model $\mu^\pattern$ of $\mT^\pattern$ in which each action $a$ is executed $\mu^R(a)$ times. Formally, if for each action $a \in A$,
$\pattern_1;a, \ldots, \pattern_k;a$ $(k \geq 1)$ are all the initial patterns of $\pattern$ ending with $a$, we have to ensure
$
\sum_{i=1}^k \mu^\pattern(a^{\pattern_i})  = \mu^R(a).$ 

Since $<$ is compatible with $\pattern$, for any  action $a$ there exists an ``$<$-compatible" action $a^{\pattern_1}$ with $\pattern_1$  an initial pattern of $\pattern$. 
$\Pi^\pattern$ dominates $\Pi^<$ because for each model $\mu^<$ of $\mT^<$ there is a model $\mu^\pattern$ of $\mT^\pattern$ assigning  $1$ to the $<$-compatible actions assigned to $\top$ by $\mu^<$, and  $0$ to the others.
\end{proof}

According to the Theorem, even restricting to simple and complete patterns $\pattern$, our pattern $\pattern$-encoding allows to find plans with a bound $n$ which is at most equal to the bound necessary when using the rolled-up, standard and \re{} $<$-encodings, the latter with $<$ compatible with $\pattern$.

\section{Implementation and Experimental Analysis}

Consider a numeric planning problem $\Pi$.
Clearly, the performances of the encoding $\Pi^\pattern$ may greatly depend on the pattern $\pattern$. 
For computing the pattern, we use  the Asymptotic Relaxed Planning Graph (ARPG) \cite{Scala_Haslum_Thiebaux_Ramirez_2016_AIBR}. An ARPG is a digraph of alternating state  ($S_i$) and action  ($A_i$)  layers, which, starting from the initial state layer, outputs a partition $A_1, \ldots, A_k$ on the set of actions which is totally ordered. 
If $a\in A_{i+1}$ then any sequence of actions which contains $a$ and which is executable in the initial state, contains at least an action  $b \in A_i$ ($0\leq i < k)$. In the computed pattern, $a$ precedes $b$ if $a \in A_i$ and $b \in A_j$ with $1 \leq i < j \leq k$, while actions in the same partition are randomly ordered.

\begin{rexample} The ARPG construction leads to the following ordered partition on the set of actions:
$\set{\ttt{lft}_r, \ttt{rgt}_r, \ttt{lft}_l, \ttt{rgt}_l, \ttt{lre}, \ttt{rle}}$, then $\set{\ttt{conn}}$, and finally $\set{\ttt{exch}, \ttt{disc} }$.
Depending on  whether  \ttt{exch} occurs before or after $\ttt{disc}$ in the pattern, the plan in equation (\ref{eq:ex:shortest-plan}) is found with bound $n=2$ or $n=3$, respectively.%
\end{rexample}

For the experimental analysis we considered  all the domains and problems of the 2023 Numeric International Planning Competition  (IPC) \cite{ipc2023}. We compared our planner \planner with the three symbolic planners 
\textsc{Springroll} (based on the rolled-up $\Pi^R$ encoding \cite{Scala_Ramirez_Haslum_Thiebaux_2016_Rolling}), a version of \planner computing the \re{} $<$-encoding $\Pi^<$ with $<$  compatible with $\pattern$, and called it \re{}; 
and \textsc{OMTPlan} (based on the $\Pi^S$ standard encoding), 
and the three search-based planners \textsc{ENHSP} \cite{Scala_Haslum_Thiebaux_Ramirez_2016_AIBR}, \textsc{MetricFF} \cite{hoffmann2003metric} and \textsc{NumericFastDownward} (\textsc{NFD}) \cite{kuroiwa2022lm}. \textsc{NFD} and \textsc{OMTPlan} are the two planners that competed in the last IPC, ranking first and second, respectively. The planner \textsc{ENHSP} has been run three times using the \texttt{sat-hadd}, \texttt{sat-hradd} and \texttt{sat-hmrphj} settings, and for each domain we report the best result we obtained \cite{Scala_Haslum_Thiebaux_2016_Subgoaling, scala2020search}.
All the symbolic planners have been run using \ttt{Z3} v4.12.2 \cite{de2008z3} for checking the satisfiability of the formula (\ref{eq:enc-sat}), represented as a set of assertions in the SMT-LIB format \cite{BarFT-SMTLIB}.
We then considered the same settings used in the Agile Track  of the IPC, and thus with a time limit of $5$ minutes. Analyses have been run on an Intel Xeon Platinum $8000$ $3.1$GHz with 8 GB of RAM.

For lack of space, Table \ref{tab:experiments} presents the results for all the planners but \textsc{OMTPlan}, since its encoding is dominated by the one of \textsc{Springroll}.%
\footnote{
The table shows the results only for those domains for which at least one planner was able to solve one problem in the domain. Our planner is available at {\url{https://pattyplan.com}}} 
In the sub-tables/columns, we show: the name of the domain (sub-table Domain); the percentage of solved instances (sub-table Coverage); the average time to find a solution, counting the time limit when the solution could not be found (sub-table Time); the average bound at which the solutions were found, computed considering the problems solved by all the symbolic planners able to solve at least one problem in the domain (sub-table Bound); the number of variables (sub-table $|\mX\cup\mA\cup\mX'|$) and assertions (sub-table $|\mT(\mX\cup\mA\cup\mX')|$) of the  encoding with bound $n=1$. For  the symbolic planners, the bound is increased starting from $n=1$ until a plan is found or resources run out. A ``-” indicates that no problem in the domain was solved by the planner with the given resources.
The table has been divided based on the average value of $|V_B|/|V_N|$: if $|V_B|/|V_N| < 1$ the domain is considered \emph{Highly Numeric} (above), and \emph{Lowly Numeric} (below) otherwise. 

From the table, considering the data about the symbolic planners in the last three sub-tables, two main observations are in order. First, \planner always finds a solution within a bound lower than or equal to the ones needed by the others (accordingly with Theorem \ref{thm:domination}). Second, even considering the bound $n=1$, \planner produces formulas with $(i)$  roughly the same number of variables as \textsc{SpringRoll} and far fewer than \re{};
and $(ii)$ (far) fewer assertions than \textsc{SpringRoll} and \re{}. %
Considering the sub-tables with the performance data,
$(i)$ on almost all the Highly Numeric domains, \planner outperforms all the planners, both symbolic and search-based; $(ii)$ in the \textsc{Drone} and \textsc{PlantWatering} domains and in all the Lowly Numeric domains, \planner outperforms all the other symbolic  planners but performs poorly wrt the search-based planners: indeed the solution for such problems usually requires a bound unreachable for \planner (and for the other symbolic planners as well); $(iii)$ 
overall, \planner{} and \textsc{ENHSP} are the planners having the highest coverage on the highest number of domains, with \planner having the best average solving time on more domains than \textsc{ENHSP} (and the other planners as well). Finally, we did some  experiments with the time-limit set to 30 minutes, 
obtaining the same overall picture.

\begin{figure}[t]
    \centering
    \includegraphics[width=\columnwidth]{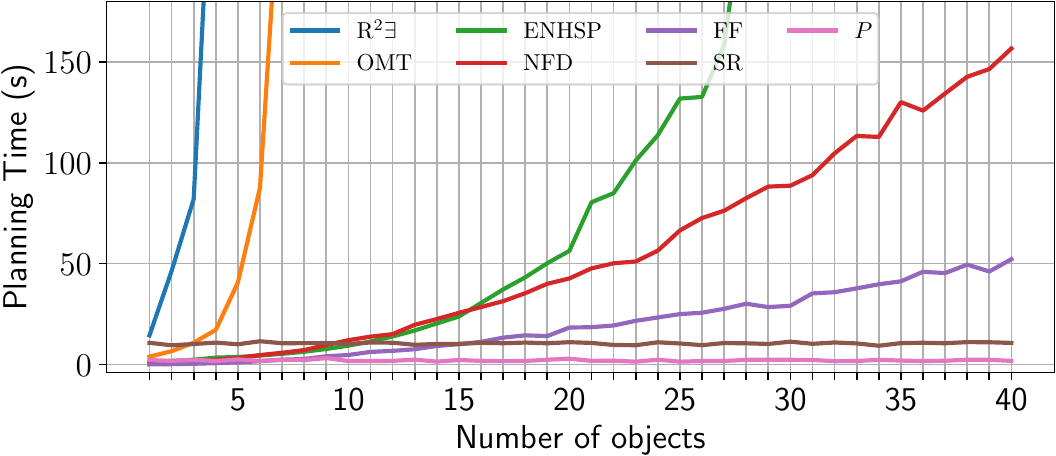}
    \caption{Performance on the \textsc{LineExchange} domain.}
    \vspace*{-5mm}
    \label{fig:plot}
\end{figure}

We also considered the \textsc{LineExchange} domain, which is a generalization of the domain in the Example. In this domain, $N=4$ robots are positioned in a line and need to exchange items while staying in their adjacent segments of length $D=2$. 
At the beginning, the first robot has $Q \in \natural$ items and  the goal is to transfer all the items to the last robot in the line. 
In Figure \ref{fig:plot}, we show how the planning time varies with $Q$: when $Q=1$, all the variables are essentially Boolean (since they have at most two possible values) and all the symbolic planners are outperformed by the search-based ones. As $Q$ increases,  rolling-up the exchange actions becomes more important, and thus  \planner and \textsc{SpringRoll} start to outperform the search-based planners. Patterns allow \planner to perform better than \textsc{SpringRoll}, while our \re{} planner performs poorly because of the high number of assertions and variables produced.

\section{Conclusions}

We  presented the pattern encoding which generalizes the state-of-the-art rolled-up and \re{} encodings. We  provided theoretical and experimental evidence of its benefits. We believe that our generalization provides a new starting point for the research in symbolic planning. Indeed, %
more research is needed to extend the ARPG construction for 
better patterns.

\bibliography{aaai24}

\end{document}


\maketitle

\section{Appendix}

\subsection{Computing the pattern via ARPG} \label{sec:arpg}

For the experimental analysis, we computed the patterns using the Asymptotic Relaxed Planning Graph (ARPG) as presented in \cite{Scala_Haslum_Thiebaux_Ramirez_2016_AIBR}. An ARPG is a digraph of alternating state layers ($S_i$) and action layers ($A_i$), In state layers, numeric variables are represented as intervals and Boolean variables as literals. In the state layer $S_0$ there are only unit intervals and literals as dictated by the initial state. In the action layer $A_i$ there are only the actions whose preconditions can be achieved by the intervals and literals in $S_i$. The state layer at $S_{i+1}$ is obtained by extending the intervals and literal of $S_{i}$ with the effects of all the actions in $A_{i}$. The layers are extended until a fix point is reached. 
The ARPG construction then
outputs a partition $A_1, \ldots, A_k$ on the set of actions which is totally ordered. In the computed pattern, $a$ precedes $b$ if $a \in A_i$ and $b \in A_j$ with $1 \leq i < j \leq k$, while two actions in the same partition are randomly ordered.

\begin{rexample} Constructing the ARPG on the example, we obtain the following layers:
\small
\begin{eqnarray*}
S_0 &=& \set{\neg p, x_l = [-X_I, -X_I], x_r = [X_I, X_I],\\ &&  q_l = [Q, Q], q_r = [0, 0], q = [1,1]},\\ 
    A_0 &=& \set{\ttt{lft}_r, \ttt{rgt}_r, \ttt{lft}_l, \ttt{rgt}_l, \ttt{lre}, \ttt{rle}},\\
    S_1 &=& \set{\neg p, x_l = [-\infty, 0], x_r = [0, +\infty], \\ && q_l = [Q_l, Q_l],  q_r = [Q_r, Q_r], q = [-1,1]}, \\
    A_1 \setminus A_0 &=& \set{\ttt{conn}}
\end{eqnarray*}
\begin{eqnarray*}
    S_2 &=& \set{\neg p, p, x_l = [-\infty, 0], x_r = [0, +\infty], \\ && q_l = [Q, Q],  q_r = [0, 0], q = [-1,1]}, \\
    A_2 \setminus A_1 &=& \set{\ttt{exch}, \ttt{disc} }\\
    S_3 &=& \set{\neg p, p, x_l = [-\infty, 0], x_r = [0, +\infty], \\ && q_l = [-\infty, +\infty],  q_r = [-\infty, +\infty], q = [-1,1]}, \\
    A_3 \setminus A_2 &=& \emptyset
\end{eqnarray*}
\normalsize
Since  \ttt{exch} and \ttt{disc} belong to the same partition, the  action \ttt{exch} can be put before or after the action $\ttt{disc}$ in the pattern, allowing for the plan in Equation (2) to be found with bound $n=2$ or $n=3$, respectively.

\end{rexample}

\bibliography{aaai23}